\documentclass[a4paper,conference]{IEEEtran}
\pdfoutput=1
\usepackage[utf8]{inputenc}
\usepackage{cite}
\usepackage{amsmath,amssymb,amsfonts}
\usepackage{algorithmic}
\usepackage{graphicx}
\usepackage{textcomp}
\usepackage{xcolor}
\usepackage{import}
\usepackage{makecell}
\usepackage{subfigure}
\usepackage{float}
\usepackage{multirow}

\def\BibTeX{{\rm B\kern-.05em{\sc i\kern-.025em b}\kern-.08em
    T\kern-.1667em\lower.7ex\hbox{E}\kern-.125emX}}
\begin{document}

\title{TimeCaps: Capturing Time Series Data With Capsule Networks\\}

\author{\IEEEauthorblockN{Hirunima Jayasekara\textsuperscript{1}, Vinoj Jayasundara\textsuperscript{1}, Mohamed Athif\textsuperscript{2}, Jathushan Rajasegaran\textsuperscript{1}, Sandaru Jayasekara\textsuperscript{1},  \\ Suranga Seneviratne\textsuperscript{3}, Ranga Rodrigo\textsuperscript{1}}
\IEEEauthorblockA{\textit{\textsuperscript{1}University of Moratuwa, Sri Lanka}\\
\IEEEauthorblockA{\textit{\textsuperscript{2}Boston University, MA, USA}\\
\IEEEauthorblockA{\textit{\textsuperscript{3}School of Computer Science, University of Sydney} \\
\{nhirunima, vinojjayasundara\}@gmail.com, athif@bu.edu,\{brjathu, sandaruamashan\}@gmail.com,\\
suranga.seneviratne@sydney.edu.au, ranga@uom.lk}}}}

\maketitle

\begin{abstract}
Capsule networks excel in understanding spatial relationships in 2D data for vision related tasks. TimeCaps is a capsule network designed to capture temporal relationships in 1D signals. In TimeCaps, we generate capsules along the temporal and channel dimensions, creating two feature detectors that learn contrasting relationships, prior to projecting the input signal in to a concise latent representation. We demonstrate the performance of TimeCaps in a variety of 1D signal processing tasks including characterisation, classification, decomposition and reconstruction. TimeCaps surpasses the state-of-the-art results by achieving 96.96\% accuracy on classifying 13 Electrocardiogram (ECG) signal beat categories, while achieving on-par results on identifying 30 classes of short audio commands. Instantiation parameters inherently learnt by the capsule networks allow  characterisation, compression and reconstruction of 1D signals. TimeCaps demonstrates to be a promising data driven technique for multiple signal processing tasks.\footnote{https://github.com/hirunima/TimeCaps
}
% Capsule networks excel in understanding spatial relationships in 2D data for vision related tasks. Even though they are not designed to capture 1D temporal relationships, with TimeCaps we demonstrate that given the ability, capsule networks excel in understanding temporal relationships. To this end, we generate capsules along the temporal and channel dimensions, creating two feature detectors which learn contrasting relationships.  TimeCaps surpasses the state-of-the-art results by achieving 96.96\% accuracy on identifying 13 Electrocardiogram (ECG) signal beat categories, while achieving on-par results on identifying 30 classes of short audio commands.
% Further, the instantiation parameters inherently learnt by the capsule networks allow us to attempt characterisation, compression and synthesis of 1D signals, which open various possibilities in signal processing.\footnote{Resources will be available upon decision}
\end{abstract}

\vspace{-2mm}
\section{Introduction}
\label{sec:introduction}
%\import{}{introduction}
Different signal processing methods are used in classification, characterisation, compression and reconstruction of time series signals. These include source separation methods such as independent component analysis, wavelet transforms or machine learning methods~\cite{yu2008integration,elhaj2016arrhythmia,kim2009robust,dixon2011compressed,kundu2015electrocardiogram,turajlic2012novel}. However, most of these signal processing methods are very specific to the application and require the user to make decisions either in the form of parameters of a deterministic model or manual feature curation in a machine learning model. 

Recently, data-driven approaches such as 1D Convolutions and Recurrent Neural Networks have produced promising results, albeit specific only for the task of classification~\cite{izci2019cardiac,acharya2017deep,mohamed2011acoustic, hinton2012deep}. In this work, we present a data-driven approach, termed TimeCaps, to facilitate multiple  applications in time series signal processing, that eliminates the necessity of `prior knowledge' about the signals of interest and in fact capable of elucidating relationships that are otherwise empirically required to formulate a model. In addition, we show that TimeCaps performs with accuracy comparable to the state-of-the-art in these applications, even with sparse data.

% \textcolor{red}{TODO Also say something on compression/reconstruction methods and limitations.}

%Suranga - To this end, we propose Timecaps, which can perform successful signal reconstruction with a very low dimensional --as low as 4 dimensions per signal-- latent representation, significantly reducing the computational complexity. TimeCaps architecture is based on the proposed Capsule Networks for image classification~cite{}

% \texttt{\textcolor{blue}{P3: Introduce the traditional usual operation of capsule networks and describe the Time Caps intuition}}

Recently, Capsule Networks~\cite{sabour2017dynamic} have been proposed to address limitations of CNNs such as the loss of spatial information in the pooling layers and being ambivalent to the spatial relationships between the learnt entities, mainly in the domain of image classification. CapsNets learn to encode the properties of an entity present in the input—in this case a signal— within the instantiation parameters, in addition to the probability of existence of the entity. Further, the capsules in one layer are \textit{dynamically routed} to the capsules in the next layer based on their agreement, formulating meaningful part-whole relationships. 
In this paper, we propose TimeCaps which adapts the ideas of 2D capsule networks to 1D signals by creating capsules along the temporal axis and along the feature map axis such that our network will capture the temporal relationships between the temporal entities. This allows us to achieve better signal encoding and classification.

% Now we should address each of the applications we are presenting, (compression and reconstruction) and (classification, characterisation)  in two separate paragraphs}

% compression
% reconstruction
Accurate compression and reconstruction of time series signals is a requirement for data storage in many domains such as medical records \cite{kundu2015electrocardiogram}, digital audio processing\cite{sturmel2011signal}.
The main drawback of  state-of-the-art reconstruction methods is that they require a high dimensional latent representation, in order to perform successful reconstruction. A majority of methods add artifacts to the reconstructed signals with dimensionality reduction\cite{dixon2011compressed,mandic2018biomedical}. TimeCaps can perform successful reconstruction with a very low dimensional --as low as four dimensions per signal-- latent representation, significantly reducing the computational complexity. 

% classification
% characterization
% \textcolor{red}{Athif: We need a few sentences like in the above paragraph  for these tasks of classification
% and characterization - we need accurate classification we need accurate characterization, example domains, why current state of the art is not good enough - why ours might be better - which is answered next} 
Signal characterization along with classification is essential when studying the discriminative features among signals in similar domains. Even though deep neural networks dominate majority of 1D signal classification tasks\cite{izci2019cardiac,mohamed2011acoustic, hinton2012deep,
graves2013speech,graves2014towards,sainath2015convolutional,
zhang2017towards,lee2017raw,sainath2015learning}, when it comes to low number of training data with imbalanced classes, they tend to under-perform compared to machine learning models.
%\cite{li2017classification}.
Yet, these models are lagging behind human-level performance. In contrast, our proposed method performs classification with a low number of training data and imbalanced classes while learning unique characterization for each signal.

Since the characteristics of the signal can be modulated by perturbing these dimensions in a class independent fashion, TimeCaps facilitates quantitative characterization of time series signals. TimeCaps learns relationships among the entities along the temporal and feature dimensions in the time series signals. Since it learns the most differentiating morphology of the time series signal from the data, it enables highly accurate classification than when features are hand-crafted.

To demonstrate the capabilities of TimeCaps, we selected two contrasting application domains: human electrocardiogram (ECG) signals and speech commands. ECG signals are one of the fundamental physiological measurements that enable the diagnosis of cardiac diseases. Pathological markers of ECG signals are encoded in the durations, amplitudes, shapes and intervals of the waveforms \cite{sornmo2005bioelectrical}. Computerised analysis of ECG can contribute towards accurate, fast, consistent and reproducible results aiding clinicians in better diagnosis. We demonstrate the suitability of TimeCaps in classification, characterisation, compression and reconstruction of ECG signals. Further, we demonstrate the robustness of TimeCaps for the classification task on a different signal domain instead of biological signals using short speech commands, resulting end-to-end trainable audio classifier.
% \textcolor{red}{Athif: Need description of this dataset like we did for the ecg in the previous sentence}.

More specifically, we make the following contributions.
\begin{itemize}
\item We introduce a new end-to-end trainable Capsule Network architecture that can simultaneously classify, encode raw 1D signals and decode the encoded signal. %which can simultaneously classify and encode raw signals as well as decode the encoded signal.
\item Our proposed network generates signal embeddings containing meaningful features of the input signal which can be directly use to identify the nature of the signal.%\textcolor{red}{ Signal embeddings derived from the proposed network contains meaningful features of the input signal which can be readily used to identify the nature of the signal}
\item To the best of our knowledge, we are the first to reconstruct a 1D signal with capsule networks.
\item We extend CapsNets' abilities to learn temporal relationships in 1D signals and explore varieties of feature maps rather than a single set of feature maps.
\item We surpassed the state-of-the-art by achieving 96.21\% accuracy on MIT-BIH ECG  Dataset\cite{moody2001impact} across 13 classes. Our model can classify rare beats which had only a few training samples. 
\item  We were able to achieve on-par classification results with the state-of-the-art on Google Speech Commands Dataset\cite{warden2018speech} on 30 classes. 
% \textcolor{blue}{Add some performance numbers e.g. accuracy?}
\end{itemize}

The rest of the paper is organized as follows. In Section \ref{sec:related_works}, we describe related work. In Section \ref{sec:method}, we explain the TimeCaps cells and the architecture, and in Section \ref{sec:results}, we present the experiments and the results. Section \ref{sec:conclusion} concludes the paper.

% \textcolor{blue}{Guys, you can leave the introduction with me.
% However, I have few questions that need to better understand what to follows.}

% \textcolor{blue}{i) It seems that we want to focus only ECG classification, However we also talk about Google speech dataset. So either we can write the intro completely focusing on ECG or we can say we provide a generic model for time series classification and we validate it with two datasets? Given the coference I am guessing it is better to write as the latter.
% }

% \textcolor{blue}{ii) Do we plan to a comparison with other methods. Table 2 seems only for ECG data? What I mean is do we plan to provide a comparison with ICA, Sparse coding or any similar method, e.g. LSTM?}

% \textcolor{blue}{iii) In the intro we need a paragraph to describe the intuition that why Capsule networks may be better for time series modeling. It have been initially proposed address limitations of CNNs mostly related to image classification. 1D Convolution on the other hand is a simple pattern matcher. So why a Capsule Network works here?   }

% \textcolor{red}{TODO: If possible let's try to have a quick conf call tomorrow.}

\vspace{-3mm}
\section{Related Work}
\label{sec:related_works}
Early work in audio classification was limited to signal processing and pattern recognition techniques like Hidden Markov Models (HMM), Gaussian Mixture Models (GMM) and Dynamic Time Warping \cite{mohri2002weighted,gales2008application} etc. Yet the capability of such models was far behind to use them in industrial applications. In the 
wake of deep learning, audio processing became mainstream with several authors using Recurrent Neural Networks (RNN), Long-Short-Term-Memory (LSTM) combined with HMM or GMM \cite{mohamed2011acoustic, hinton2012deep} and some focus on end-to-end training \cite{graves2013speech,graves2014towards}. Alternatively, some tend to use CNNs with Mel-frequency cepstrum coefficients (MFCC)\cite{sainath2015convolutional} as the feature extractor for the CNN. However, there are a few works\cite{lee2017raw,sainath2015learning} that have been done for learning features of the raw signal while preserving the end-to-end trainability of the network.

% About ecg beat classification
Electrocardiogram (ECG) signal analysis plays a vital role in medical diagnosis since ECG signal can provide vital information that can help to diagnose various health conditions. For example, ECG beat classification; classifying ECG signal portions into classes such as normal beats or different  arrhythmia types such as atrial fibrillation, premature contraction, or ventricular fibrillation allows to identify different cardiovascular diseases. 

One of the early ideas of ECG is Independent Component Analysis (ICA) that dates back to 1996 and subsequently used ICA on EEG signals~\cite{makeig1996independent}. Following the success, several authors used ICA on ECG signals ~\cite{yu2008integration,elhaj2016arrhythmia} as a feature extracting mechanism to train different classifiers.
Progressive learning of suitable features offered by deep learning based approaches have proven to outperform hand crafted feature-based approaches in a wide range of applications, including bio-medical signal processing for classification ~\cite{yu2008integration,elhaj2016arrhythmia,kim2009robust} and reconstruction tasks~\cite{dixon2011compressed,mandic2018biomedical}.   
% Due to high precision rates and convenience, deep learning based models were able to achieve the state-of-the-art results for many areas. 
For an instance, MIT-BIH Arrhythmia Database ~\cite{moody2001impact} is a widely used ECG signal dataset to characterize, classify and generate ECG beats corresponding to heart diseases.
% Due to the unavailability of high number of data samples for each class, most classification methods only consider a sub-set of the full dataset\cite{martis2013ecg,kim2009robust}.
Similarly,  ECG signal compression and reconstruction have a variety of applications such as remote cardiac monitoring in body sensor nodes~\cite{mamaghanian2011compressed} and achieving low power consumption when sending and processing data through IoT-gateways~\cite{al2018ecg}.
% About ecg beat reconstruction
Different approaches exist in the literature for reconstructing bio-medical signals, including Compressed Sensing (CS) ~\cite{dixon2011compressed,mandic2018biomedical} and reconstructing corrupt or missing intervals of ECG signals ~\cite{martin2014ecg}. One drawback is CS-based methods do not offer end-to-end compatibility when reconstructing a signal. Further, the capsule learnt latent parameters can be easily explainable by a simple perturbation. Moreover, CS-based methods require a higher dimensional latent representation to reconstruct a single beat, whereas our approach can reconstruct a single beat from as low as a 4-dimensional latent representation.

Traditional signal processing methods were highly affected by the wake of deep learning. Especially, several authors~\cite{izci2019cardiac,acharya2017deep} suggested CNNs for classifying ECG signals. The CNNs in both approaches learned the features of the signal using 1D convolutional kernels. Even though CNN architectures achieve state-of-the-art classification results, they consist of several drawbacks. CNN models disregard the spatial relationship in input data while needing thousands of data points to achieve a good performance. Capsule networks (CapsNet)~\cite{sabour2017dynamic}, in contrast, solves this problem by learning the properties of an entity in addition to its existence. First step towards the idea of capsule network was introduced in transforming auto encoders~\cite{hinton2011transforming}. Later Sabour \emph{et al.}~\cite{sabour2017dynamic} proposed dynamic routing between capsules and able to achieve on-par results with state-of-the-art CNN models. However, a vanilla capsule network consists of only two convolution layers and a fully connected dynamic routing layer. In order to go deeper with CapsNet, Rajasegaran \emph{et al.}~\cite{rajasegaran2019deepcaps} have suggested a new capsule layer which shares parameters across capsules. Incorporating time series data with capsule network was introduced by ~\cite{bae2018end}, yet, input to the model was hand crafted MFCCs. Further, the authors do not utilize the decoder network to decode the signal hence they were unable to discuss the use of instantiation parameters in their work.

Few recent works utilize the 1D version of capsule networks. Zhang \emph{et al.}~\cite{zhang20191d} used 1D capsule networks for the hyper-spectral image classification by applying 1D convolution in separate spatial and spectral domains, to reduce the number of parameters. Berman \emph{et al.}~\cite{berman2019dga} used 1D capsules for domain generation algorithm detection, by learning the features with a single dimension and passing them directly to a LSTM learner. Further, Butun \emph{et al.}~\cite{butun20201d} applied 1D convolution followed by 1D routing to the ECG signals. This is the closest to our work, yet, in our method, we consider temporal and spatial features separately and apply dynamic routing in different directions.

\vspace{-1mm}
\section{TimeCaps}
\label{sec:method}
Capsule networks inherently learn unique properties of the input images, such as location, scale, rotation and luminosity. Sabour \textit{et al.}~\cite{sabour2017dynamic} proposed a capsule architecture, to learn key features from images for classification. However, the concept of dynamic routing is not limited to images or any other form of 2D signals. 1D signals are one such example, albeit consisting of a different kind of feature hierarchy than 2D images. In this paper, we design a capsule inspired dynamic routing algorithm, which inherently learns the temporal relationships in continuous signals.

In TimeCaps, we treat the input as time series data, $X_{sig}\in \mathbb{R}^L$ and $L$ is the length of the signal. Hence, rather than learning the spatial relationship between incoming capsules, we learn the temporal relationships between adjacent capsules. These relationships are learnt by predicting the past and future capsule outputs for a given capsule.

First, we transform the input from signal domain to feature space by convolving the input signal $X_{sig}$ with $k$ number of $\psi_i$ kernels, where $\psi_i \in \mathbb{R}^{g_1}$, $i\in [1,k]$ and $g_1$ is the filter size, resulting in $\Phi_{conv1}\in \mathbb{R}^{(L\times K)}$. We pad every convolution along the temporal dimension appropriately to maintain it at size $L$ for consistency, especially during the reshaping operations performed in steps \ref{sec:timecapsA} and \ref{sec:timecapsB}. Subsequently, $\Phi_{conv1}$ will be fed to the two parallel modules TimeCaps Cell A and TimeCaps Cell B.

\begin{figure}[ht]
  \centering
  \includegraphics[scale=0.62
  ]{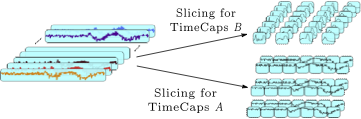}  
  \caption{Generating capsules along temporal axis and channel axis to act as temporal feature detectors.}
 \label{fig:timecaps_slice}
 \vspace{-5mm}
\end{figure}

When exploring the feature detectors, we intend to learn different types of features, and therefore we slice along the temporal axis as well as the feature axis to produce two different capsule representations called TimeCaps $\mathit{A}$ and TimeCaps $\mathit{B}$ as illustrated in Fig. \ref{fig:timecaps_slice}. Here, TimeCapsule A contains the local temporal information, while TimeCapsule B encapsulates local spatial relationships. This help us to separably learn two contrasting types of relationships and able to fuse them latter stages.

% \begin{figure}[!h]
%   \centering
%   \includegraphics[angle=-90, scale=0.69]{fig/1.pdf}  %\setlength{\belowcaptionskip}{-1cm}
%   \caption{Generating capsules along temporal axis and channel axis to act as temporal feature detectors.}
%  \label{fig:timecaps_slice}
% \end{figure}
\subsection{TimeCaps Cell $\mathit{A}$} \label{sec:timecapsA}
Timecaps A keeps the samples into one single frame, but it tries to predict the possible feature maps in the next layer.
First, the input to the TimeCaps Cell \textit{A}, $\Phi_{conv1}$, is convolved with $(c^p\times a^p)$ number of $\psi_m$ kernels, forming $\Phi_{conv}^A \in \mathbb{R}^{L\times (c^p\times a^p)}$, where $\psi_m \in \mathbb{R}^{g_2}$, $m\in [1,c^p\times a^p]$, $c^p$ and $a^p$ are the number of channels and the dimensionality of the subsequently formed primary time capsules respectively. To facilitate the formation of capsules by bundling sets of feature maps together, $\Phi_{conv}^A$ is reshaped in to $\Bar{\Phi}_{conv}^A\in \mathbb{R}^{L\times C^p\times a^p}$ and $\mathit{squashed}$ along the feature axis \cite{sabour2017dynamic} to create output of the Primary Time Capsules \textit{A}, $\Omega_{PTC}^A\in \mathbb{R}^{L\times C^p\times a^p}$.

Instead of using a traditional transformation matrix to transform low dimensional features to high dimensional features as suggested by \cite{sabour2017dynamic} we used a convolution kernel which can be used to predict the next set of capsules for the given capsule. These predicted capsules are called votes for the TimeCaps. When considering the TimeCaps $\mathit{A}$, predicted capsules are corresponds to the bundle of high feature maps.

In order to be compatible with with 2D convolution, we reshape $\Omega_{PTC}^A$ into $\bar{\Omega}_{PTC}^A$ with shape $L\times (c^p\times a^p) \times 1$. By convolving $\bar{\Omega}_{PTC}^A$ with $\psi_m^A \in \mathbb{R}^{g_3\times a^p}$, $m\in [1,c^{SA}\times a^{SA}]$ with strides $[1,a^p]$, we generate the votes $W^A_{conv}$ with shape $L \times c^p\times(c^{SA}\times a^{SA})$, for the capsules in the subsequent time capsule layer. Here, $c^{TA}$ and $a^{TA}$ are the number of channels and the dimensionality of the time capsules respectively. In consistence with the previous convolutions, the temporal dimension is kept at size $L$, whereas the size of the channel dimension is calculated by $\frac{c^p\times a^p - a^p +0}{a^p} + 1 = c^p$.
% \begin{equation}
%   \frac{c^p\times a^p - a^p +0}{a^p} + 1 = c^p
%     \label{eq:stride}
% \end{equation}
% then TimeCaps layer produce votes $W^A_{conv} \in \mathbb{R}^{L \times C^p\times(C^{SA}\times a^{SA})}$ for the next layer by convolving $\bar{\Omega}_{PTC}^A$ with $\psi_m^A \in \mathbb{R}^{g\times a^p}$ ,$m\in [1,C^{SA}\times a^{SA}]$ with strides $a^p$ as given in Eq. \ref{eq:stride}. 
% \begin{equation}
%   \frac{C^p\times a^p - a^p +0}{a^p}=C^p
%     \label{eq:stride}
% \end{equation}
To facilitate dynamic routing,  we reshape the votes $W^A_{conv}$ to $\bar{W}^A_{conv}$, to have the shape $(L \times c^p\times c^{SA}\times a^{SA})$. Subsequently, we feed the modified votes to the routing algorithm which is described in section \ref{sec:routing}. The resulting tensor, $\bar{\Omega}_{STC}^A\in \mathbb{R}^{L\times C^{SA}\times a^{SA}}$ will be flattened along the first two axes of the tensor while keep the last axes (dimension of secondary capsules) constant generating set of flattened capsules $\Omega^A\in \mathbb{R}^{(L\times C^{SA})\times a^{SA}}$ as the output of TimesCaps Cell A.
\subsection{TimeCaps Cell $\mathit{B}$} \label{sec:timecapsB}
Different from TimeCaps Cell A, Timecaps Cell B is designed to predict future and past values of a small segment of the signal. This will require to $\mathit{squash}$ along the full set of feature maps, first we reduce the number of feature maps by performing $1\times 1$ convolutions abreast to the first convolution layer of TimeCaps Cell B to create $C^b\times a^b$ feature maps. Then complementing the above idea, primary capsules were created by segmenting the final convolution output $\Phi_{conv}^B \in \mathbb{R}^{L\times (C^b\times a^b)}$ into $n$ size segments resulting $\bar{\Phi}_{conv}^B \in \mathbb{R}^{\frac{L}{n}\times n\times (C^b\times a^b)}$ then similar to TimeCaps Cell A, $\bar{\Phi}_{conv}^B$ then be $\mathit{squash}$ and reshaped into $\bar{\Omega}_{PTC}^B\in \mathbb{R}^{\frac{L}{n}\times (n\times C^b\times a^b) \times 1}$ to form the input to the TimeCaps B. Then votes $W^B_{conv} \in \mathbb{R}^{\frac{L}{n} \times n\times(C^{SB}\times a^{SB})}$  corresponds to the Cell B is derived by convoluting $\bar{\Omega}_{STC}^B$ with kernel $\psi_k^B \in \mathbb{R}^{g\times (C^b \times a^b)}$ ,$k\in [1,C^{SB}\times a^{SB}]$ with strides $C^b \times a^b)$. Then analogous to TimeCaps A, reshaped  $\bar{W}^B_{conv} \in \mathbb{R}^{\frac{L}{n} \times C^{SB}\times a^{SB}\times n}$ is utilized as the votes for the TimeCaps layer. These prediction capsules are corresponding to the future and past time segments of a given capsule which contains information about a given time segment. This will be routed producing $\bar{\Omega}_{STC}^B\in \mathbb{R}^{\frac{L}{n}\times C^{SB}\times a^{SB}}$. After flattening, final output of the TimeCaps Cell B would be $\Omega^B\in \mathbb{R}^{(\frac{L}{n}\times C^{SB})\times a^{SB}}$.
\vspace{-2mm}
\subsection{Routing}
\label{sec:routing}
Let votes be $V\in \mathbb{R}^{(L^l,w^l,w^{l+1},n^{l+1})}$ for the routing. Then we route a block of capsules $s$ from the child capsule to the parent capsule. 

During the routing, the coupling coefficients for each block of capsules $K_s$ are generated by applying the $\mathit{softmax}$ function on logits $B_s$ (initialized as 0) as given by Eq. \ref{eq:softmax}.
\begin{equation}
    k_{prs}=\frac{exp(b_{prs})}{\sum_x\sum_y exp(b_{prs})}
    \label{eq:softmax}
\end{equation}
Calculated $k_{prs}$ where $p \in L^{l+1} , r\in $ will be used to weigh the predictions $V_{prs}$ to get a single prediction $S_{pr}$ as given in Eq. \ref{eq:weight}, followed by a $\mathit{squash}$ function to produce $\hat{S}_{pr}$. $\mathit{squash}$ is used to suppress the low probabilities and to enhance the high probabilities in the prediction vectors.
\begin{equation}
b_{prs}\gets \sum_s k_{prs}\cdot V_{prs}
    \label{eq:weight}
\end{equation}
Amount of agreement between $S$ and $V$, can be measured by taking the dot product of the tensors and this will be used to update the logits in the next iteration of the routing as given in Eq. \ref{eq:update}
\begin{equation}
    b_{prs}\gets b_{prs}+ S_{prs}\cdot V_{prs}
    \label{eq:update}
\end{equation}

\vspace{-5mm}
% Each input signal contains one full ECG beat with 360 data points. In order to extract the low level features of the signal, first layer of the network are 1D convolution layer with 64 number of 1$\times$9 . 
% Next layer preprocessed the data for the timecaps layer.
% Let us have $\Phi \in \mathbb{R}^{t \times 1}$ time series frame into the model.
% \subsection{TimeCaps A}
% This layer produce the capsules by combining $u$ number of feature maps together. resulting $n$ number of snippets from one frame. One snippet $\Phi_j \in \mathbb{R}^{u \times 1}, u \times n = t$ will contains $u$ number of samples from the original signal. \vj{go with num. slices here}
% Then the preprocessing layer of layer $\mathbf{B}$ keep the samples into one single frame, but it tries to predict the possible feature maps in the next layer. Therefore, predictable feature maps will be bundle together.
% \subsection{TimeCaps B}
% As the third layer, we have two parallel timecaps layers.The preprocessing layer of the layer $\mathbf{A}$ segments the input frames into $n$ time frames (\textit{eg.} $n=10$). Then inside the Time Capsule layer, each frame will try to predict it's future and present values.
\vspace{-2mm}
\subsection{Concatenation Layer}
Since each layer explores different temporal relationships in the signal, we used a concatenation layer to concatenate the two flattened Timecaps together along the capsule axis.

Let $\alpha,\beta$ be two learnable parameters, 
Then Concatenation output would be  
\begin{equation}
    \Omega_{CC}=concat(\alpha \Omega_A,\beta \Omega_B)
    \end{equation}
Where\footnote{To facilitate concatenation, its should be held that $a^{SA}=a^{SB}=a^S$} $\Omega_A \in \mathbb{R}^{(L\times C^{SA})\times a^{SA}}$ , 
$\Omega_B \in\mathbb{R}^{(\frac{L}{n}\times C^{SB})\times a^{SB}}$ and   $\Omega_{CC} \in\mathbb{R}^{N\times a^S}$, $N=L\times C^{SA} + \frac{L}{n}\times C^{SB}$s

% Contribution ratio of each timecap layer will be analyzed in the section xx.
% \vj{Include $\alpha, \beta$ empirical results and compare with the hypothesis}

% Effect on $\alpha$ and $\beta$ on the network will be analysized in Section~\ref{sec:classification} 
\vspace{-2mm}
\subsection{Classification Layer}
As the final layer we adopt the classification layer proposed by \cite{sabour2017dynamic} to produce instantiation parameter vector $\Omega_{sig} \in \mathbb{R}^{1\times a^{sig}}$ corresponding to the signal. Class probability can be derived from the length of the vector $\Omega_{sig}$.

\begin{figure}[H]
  \centering
  \includegraphics[scale=0.36]{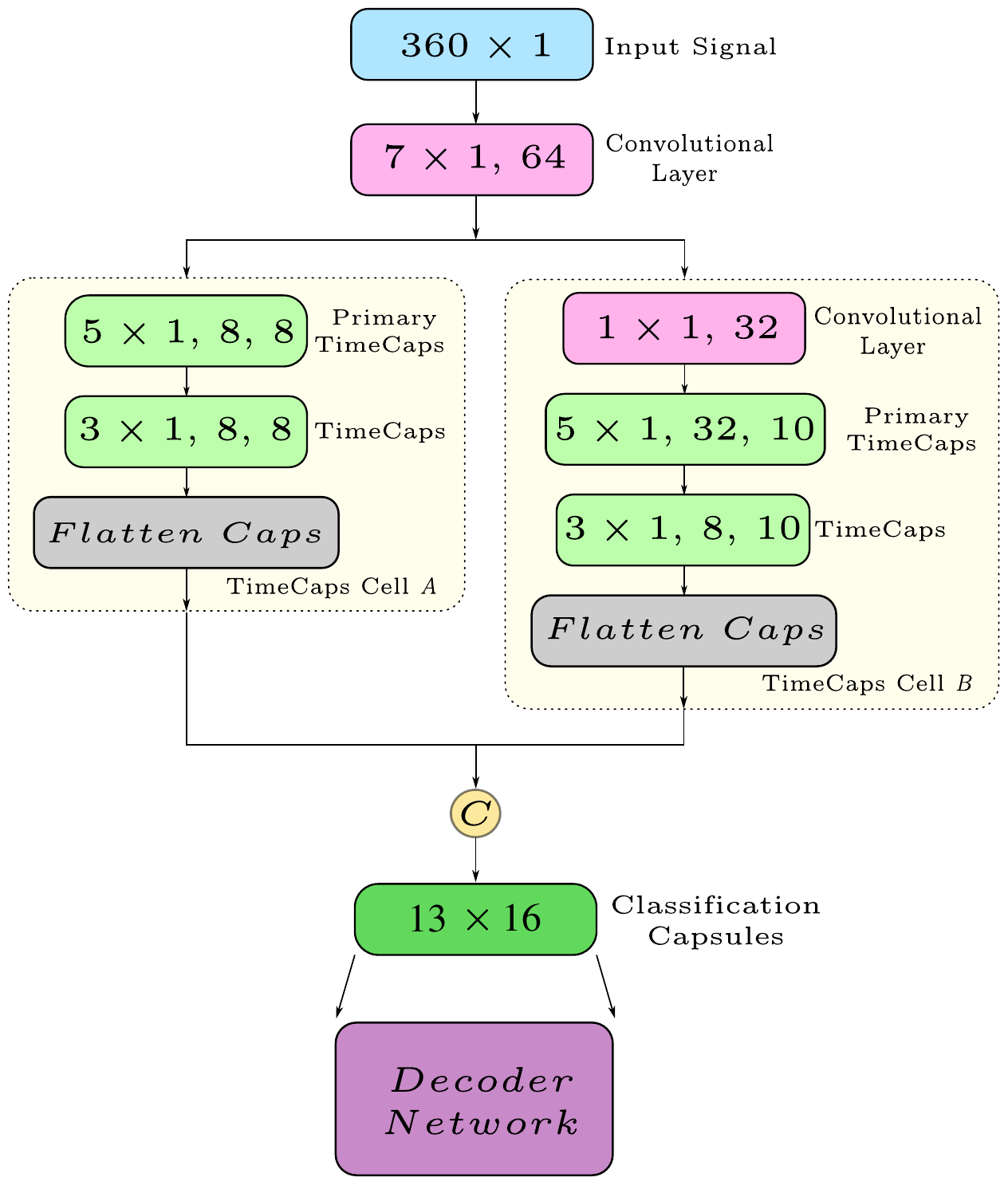}  %\setlength{\belowcaptionskip}{-1cm}
  \caption{\textbf{TimeCaps Model}: Proposed Timecaps model for time series data. TimeCaps cells \textit{A} and \textit{B} are concatenated after weighting with trainable scalars $\alpha$ and $\beta$ respectively.}
 \label{fig:timecaps_net}
 \vspace{-2mm}
\end{figure}

Full network with numerical values is illustrated in Fig. \ref{fig:timecaps_net} 
% \begin{figure}[!h]
%   \centering
%   \includegraphics[scale=0.36]{fig/network_full.eps}  %\setlength{\belowcaptionskip}{-1cm}
%   \caption{\textbf{TimeCaps Model}: Proposed Timecaps model for time series data. TimeCaps cells \textit{A} and \textit{B} are concatenated after weighting with trainable scalars $\alpha$ and $\beta$ respectively.}
%  \label{fig:timecaps_net}
% \end{figure}
% \vspace{1mm}

\subsection{Decoder Network}
\begin{figure}[ht]
  \centering
  \includegraphics[scale=0.57,angle=-90]{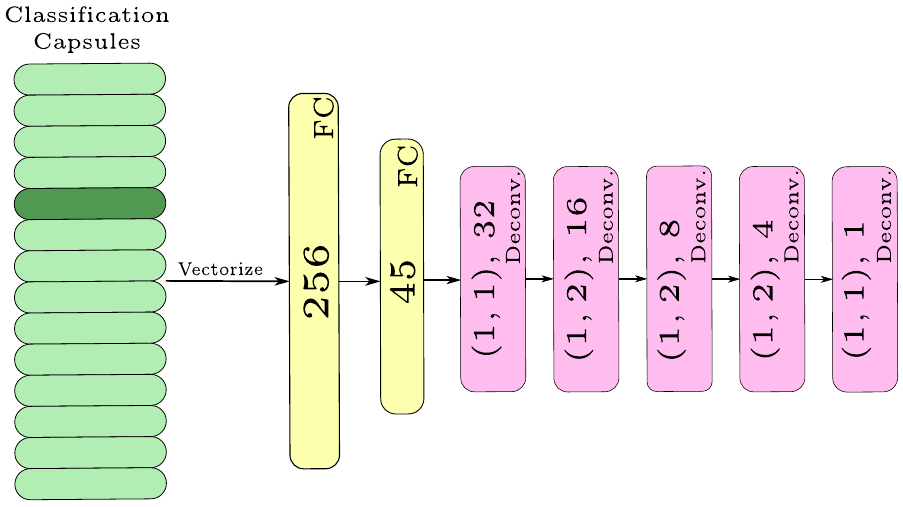}  %\setlength{\belowcaptionskip}{-1cm}
  \caption{\textbf{TimeCaps Decoder}: Decoder network for signal reconstruction. Input to the decoder network is obtained by masking the output of classification capsules.}
 \label{fig:decoder}
\end{figure}

% \vspace{-2mm}
A Decoder Network is used to reconstruct the input signal from the instantiation parameter vector $\Omega_{sig}$ extracted at the $\mathit{classification \_layer}$. Further, the decoder network provides a regularization to the Timecaps network. To this end, several papers have used this technique to reconstruct 2D image data using fully connected networks\cite{sabour2017dynamic} or 2D deconvolution networks\cite{rajasegaran2019deepcaps,jayasundara2019textcaps} which are more suitable to reconstruct image data. In this paper we applied 1D deconvolution network to decode the instantiation parameters to reconstruct 1D signals. The first two layers of the decoder are fully connected layers, followed by five 1D deconvolution layers as given in Fig. \ref{fig:decoder}. 

The existing decoder extracts the output from the $\mathit{classification \_layer}$ and masks the instantiation parameter matrix with zeros except for the row which corresponds to the predicted class. Thus, the decoder class dependable and variations causes by each instantiation parameter becomes inconsistent for each class. Our hypothesis was to construct a network such a way that the network learns to identify the fundamental features of the dataset. Therefore, instantiation parameters should be distributed  in the same space regardless of the class. As a result, we adopt a class independent decoder where prior to passing the output from the $\mathit{classification \_layer}$, we extract the $\Omega_{sig}$ of predicted class. For training, $\Omega_{sig}$ can be computed using the ground truth label while during the inference, $\Omega_{sig}$ corresponds to the class which has the maximum class probability.

% \begin{figure}[!h]
%   \centering
%   \includegraphics[scale=0.5,angle=90]{fig/decoder.eps}  %\setlength{\belowcaptionskip}{-1cm}
%   \caption{\textbf{TimeCaps Decoder}: Decoder network for signal reconstruction. Input to the decoder network is obtained by masking the output of classification capsules.}
%  \label{fig:decoder}
% \end{figure}

\subsection{Loss Function}

For the classification, we used marginal loss \cite{sabour2017dynamic} which suppresses the probabilities of other classes while enhancing the probability of true class.
\begin{equation}
\begin{aligned}
\label{eq:loss}
L_k &= T_k \max(0, m^+ - \|v_k\|)^2 \\
    &\ + \lambda (1-T_k)\max(0, \|v_k\| - m^-)^2 \\
\end{aligned}
\end{equation}

Here $L_k$ denotes the marginal loss for the class $k$ and $v_k$ is the respective output from the final capsule layer. Lower bound and the upper bound are set to $m^+ = 0.9$ and $m^- = 0.1$. $T_k$ will be set to 1 if $k$ is the true class and zero otherwise.
Similarly, for the decoder we used $MSE$ loss as suggested by \cite{sabour2017dynamic}.
\subsection{Signal Perturbation}
One of the main features of capsule networks was the ability to reconstruct input signal solely using the extracted instantiation parameters using a trained decoder network\cite{sabour2017dynamic}. In order to analyse the effect of each instantiation parameter variation on the reconstruction signal, we can add a noise vector $N_{purtb} \in \mathbb{R}^{1\times a^{sig}}$ to the extracted instantiation vector and reconstruct the purturb vector $\Theta_{purtb} \in \mathbb{R}^{1\times a^{sig}}$ using the pre-trained decoder network.
\begin{equation}
{\Theta_{purtb}}_{i}={\Omega_{sig}}_{i} + {N_{purtb}}_{i} 
    \end{equation}
Where $i \in \{0,dim(\Omega_{sig})\}$; 
We hypothesis that, similar to \cite{jayasundara2019textcaps}, by introducing a controlled noise will generate ample versions of input data sample. Yet, our objective is to analyze the variations of the input signal and not to generate new singles. Therefore, we manually picked the noise vector between [-1,1].

\section{Experiments and Results}
\label{sec:results}
We trained the TimeCaps on MIT-BIH Arrhythmia Dataset \cite{moody2001impact} which contains 16 classes of Electrocardiography (ECG) beats. We used wfdb software package\footnote{https://archive.physionet.org/physiotools/wfdb.shtml} to segment each ECG beat. Moreover, to test the applicability of TimeCaps in other domains, we used  the Google speech commands dataset \cite{warden2018speech} which contained 30 different audio commands. we discover that one-to-one comparison of all aspects of our method with the state-of-the art is impractical. Therefore, when evaluating the results, we could only compare the classification accuracy with the state-of-the-art methods for the given datasets.
% The dataset was readily available with accurate annotations per beat.
\begin{table}[!h]
\caption{Dataset statistics}
\label{table:dataset}
\centering
\footnotesize
\begin{tabular}{|>{\centering\arraybackslash}m{2cm}|>{\centering\arraybackslash}m{1cm}|>{\centering\arraybackslash}m{0.9cm}|>{\centering\arraybackslash}m{0.9cm}|>{\centering\arraybackslash}m{0.9cm}|} 
\hline
Dataset & Number of Classes  &  Train size & Test size&Sample length\\
\hline
MIT-BIH & 16 & 12963 & 4329 & 360\\ 
Google speech commands&30&51776&12944&16000\\
\hline
\end{tabular}
\vspace{-5mm}
\end{table}
\subsection{Implementation}
\label{sec:classification}
% \vj{Include the ablation study}
We used Keras and Tensorflow libraries for the development. We trained the network for 50 epochs and we used Adam optimizer with an initial learning rate of 0.001 and $\lambda$ defined in Eq. \ref{eq:loss} set to 0.5. The models
were trained on GTX-1080 and V100 GPUs, 
%and a weighted average ensembling was used for the 7-ensemble models reported in Table 1.

%Suranga - Add a footnote to th wfdb package
%Suranga - 35 epochsseems too small to me. Were you using some pre-trained model?
%Suranga - Table 1 Still contains Google Dataset?
\subsection{TimeCaps on ECG Signals}
Due to to the class imbalance present in the dataset, researchers have either used the AAMI standard to fuse sets of classes to produce five groups (N,S,V,F,Q) \cite{izci2019cardiac,elhaj2016arrhythmia} or use a subset of classes \cite{rajkumar2019arrhythmia}. We followed the latter, and used 13 beat classes (only beat types(N,L,R,B,A,a,J,V,F,j,E,/,f))\footnote{https://archive.physionet.org/physiobank/database/html/mitdbdir/intro.htm} out of 16 beat classes for classification, since the rest of the classes had lower than 50 training samples per class.

Our model consists of 6,149,856 number of trainable parameters. Table \ref{table:results_ecg} compares our results to the state of the art. However, all the work listed in Table 2 did not provide enough resources to reproduce their results on the whole dataset. Hence, we followed the same approach as the authors of\cite{izci2019cardiac}, resulting in the comparison illustrated in Table 2.

% \begin{table}[!h]
% \caption{Comparison of TimeCaps with state-of-the-art results}
% \label{table:results_ecg}
% \centering
% \footnotesize
% \begin{tabular}{|>{\centering\arraybackslash}m{2.7cm}|>{\centering\arraybackslash}m{1.8cm}|>{\centering\arraybackslash}m{1.6cm}|} 
% \hline
% Implementation & Number of Classes  &  Accuracy\\
% \hline
% Yu \textit{et al}~\cite{yu2008integration}&8& 98.00\%\\ 
% \hline
% Martis \textit{et al}~\cite{martis2013ecg}&5&99.28\%\\
% \hline
% Kim \textit{et al}~\cite{kim2009robust}&6&99.50\%\\
% \hline
% Li \textit{et al}~\cite{li2017classification}&5&97.50\%\\
% \hline
% \textbf{TimeCaps}&\textbf{13} & \textbf{96.96\%}\\
% \hline
% \textbf{TimeCaps}&\textbf{5} & \textbf{97.94\%}\\
% \hline
% \end{tabular}
% \end{table}

\begin{table}[!h]
\vspace{-2mm}
\caption{Comparison of TimeCaps with state-of-the-art results}
\label{table:results_ecg}
\centering
\footnotesize
\begin{tabular}{|>{\centering\arraybackslash}m{2.3cm}|>{\centering\arraybackslash}m{0.75cm}|>{\centering\arraybackslash}m{0.9cm}|>{\centering\arraybackslash}m{3.1cm}|} 
\hline
Implementation & Number of Classes &  Accuracy & Additional Information \\
\hline
\hline
% Yu \textit{et al}~\cite{yu2008integration}&8& 98.00\%&\\ 
% \hline
Elhaj \textit{et al}~\cite{elhaj2016arrhythmia}&5&98.91\%&AAMI classes (N,S,V,F,Q)\\
Izci \textit{et al}~\cite{izci2019cardiac}&5&97.42\%&AAMI classes (N,S,V,F,Q)\\
Rajkumar \textit{et al}~\cite{rajkumar2019arrhythmia}&7&93.6\%&(V,/,L,R,!,A,N)\\
\hline
\textbf{TimeCaps}&\textbf{13} & \textbf{96.96\%}&All the beat types\\
% \hline
\textbf{TimeCaps}&\textbf{5} & \textbf{97.94}\%&AAMI classes (N,S,V,F,Q)\\
\hline
\end{tabular}
\vspace{-2mm}
\end{table}

Due to low number of training samples present in set of classes in MIT-BIH data set, current state-of-the-art for the classification according to AAMI standard, with only 5 number of infusion classes is $98.91\%$ which was achieved by Elhaj \textit{et al} \cite{elhaj2016arrhythmia}, whereas we achieved $96.21\%$ accuracy for 13 classes. Proving Timecaps ability to work with very low number of training samples. In addition, we trained our model with the AAMI dataset, obtained an accuracy of $97.94\%$, which is on par with the state-of-the-art.

% \begin{figure}[!h]
%   \centering
%   \includegraphics[scale=0.3]{fig/cm.eps}  %\setlength{\belowcaptionskip}{-1cm}
%   \caption{Normalized confusion matrix for the ECG beat classifier. Each symbol denote a beat type\protect\footnotemark}
%  \label{fig:cm}

% \end{figure}

Since TimeCaps is based on capsule network, it inherits the high computational complexity. In TimeCaps, model complexity is heavily dependent on the number of classes in the dataset, since this directly affects the number of parameters and routing in the classification capsules.  Further, choosing the number of segments in the temporal axis (n) for TimeCaps B will depend on the nature of the signal as well as the length of the input signal.

Further, Timecaps possess the ability of encoding each beat with low dimensional vector and reconstructing each signal with the decoder network. 

Moreover, we observed that convergence of the network with a weighted concatenation of different features is much slower than 1:1 concatenation. 
\subsection{TimeCaps on Audio Signals}
% Current state-of-the-art for audio signal classification is highly influence by deep learning based models. Several authors(?;?;?;?) have used CNN based models on the spectrum of the signal. Widely used spectrum generation technique is Mel Frequency Cepstral Coefficients(MFCC)() which represent the short term power spectrum of a sound. There are several attempts() to remove the use of MFCC as a feature extractor and train a model which can identify dataset specific features by learning.
In order to explore the robustness of the network, we test our proposed architecture with Google Speech Commands Dataset~\cite{warden2018speech} which contains one second long audio commands. Even though the common practice is to extract audio features using Mel-frequency cepstral coefficients (MFCC) and feeding the extracted features to the network, some work in literature explore the ability to classifying raw audio signals \cite{lee2017raw} without pre-defining set of feature extractors such as MFCC. We adopt the latter due to the performance gain \cite{sainath2015learning} and to avoid the additional computational overhead.

% \textcolor{blue}{Athif: Do we need to justify why we dis not use MFCC - as in is it a positive thing or negative thing that we did not use, or is it neutral, if so we can just change the tone to not sound like we are giving an excuse for not using it}

% \textcolor{red}{
% When tuning the TimeCaps for audio signals, in order to facilitate a full audio command into the network the network needed to be adjusted accordingly. The main challenge was requirement of high computational complexity at TimeCaps Cells due to lengthy input and sequentially at classification Capsules due to large number of capsules from previous layer as well as comparatively high number of dataset classes. Since authors of \cite{jayasundara2019textcaps} have utilized capsule networks with 42 classes, we are not intend to address the latter and we have introduced small modifications as given in Fig. \ref{fig:audio_conv} which significantly reduce the number of parameters at TimeCaps Cells. 
% }

\begin{figure}[!h]
\vspace{-1mm}
  \centering
  \includegraphics[scale=0.4,angle=90]{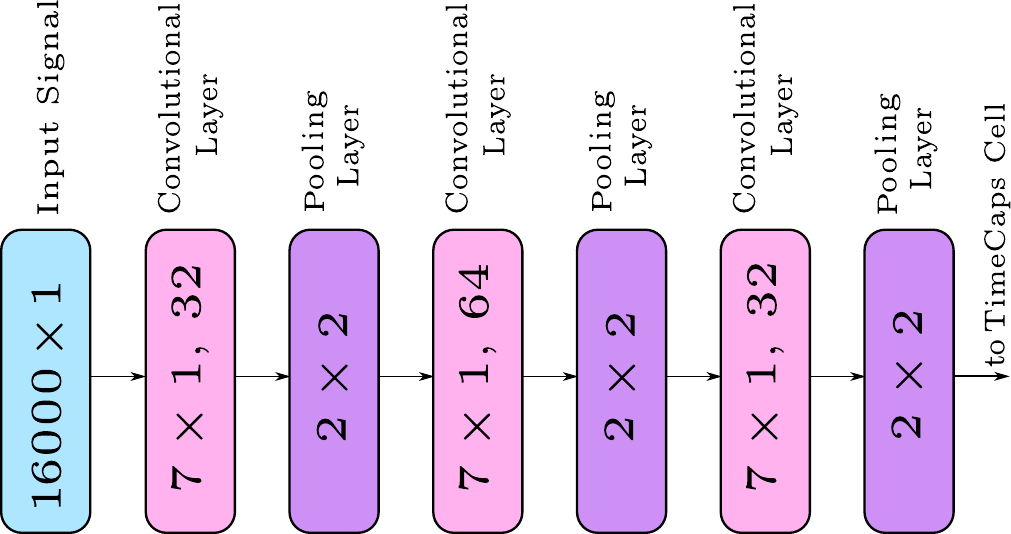}  %\setlength{\belowcaptionskip}{-1cm}
  \caption{Timecaps model adjusted for audio commands. Size of the feature maps are decreased using pooling layers before feeding to the capsule layers}
 \label{fig:audio_conv}
\end{figure}

In order to input a full audio command into TimeCaps, the network requires several key modifications. The high  sampling rate of the audio signals poses a key challenge by significantly surging the computational complexity at the TimeCaps cell, and subsequently producing a larger number of capsules to be sequentially passed down the network. Besides the computational complexity, the large number of capsules inherently degrades the performance, when the number of capsules increase, the coupling coefficients of the dynamic routing becomes less effective. This may be viewed as a theoretical issue as well of the dynamic routing algorithm\cite{xi2017capsule}. Addressing this issue is beyond the scope of this study.
As a solution to the high computational complexity, we propose the modifications proposed in Fig. \ref{fig:audio_conv}, which significantly reduce the number of parameters at TimeCaps Cells.
% \textcolor{red}{
% In order to reduce the feature map size at early stages,
% % extracted the feature maps from the first convolution layer undergo thru a $2\times2$ pooling layer and the
% input signal goes through a three sequential blocks(a block consists of convolutional layer and a $2\times2$ pooling layer) to extract the low level features of the audio command. These extracted feature maps then directed to TimeCaps Cells similar to Sec \ref{sec:method} 
% }
In order to reduce size of the feature maps at early stages, the input is passed through three sequential blocks (each block consists of a convolutional layer followed by a $2\times2$ pooling layer) to extract the low level features of the audio command. These extracted feature maps are then directed to TimeCaps Cells similar to Sec \ref{sec:method}. However, with the high dimensionality (16,000+) of the audio signals, its hard to train a decoder jointly with an encoder.

Table \ref{table:results_audio} compares our results to the state of the art with the proposed method. Even though our results are slightly below with the state-of-the-art results, our results surpass the existing capsule network model demonstrating 3.15\% boost due to the improvement provided by the TimeCaps.
\begin{table}[!h]
\caption{Comparison of TimeCaps with state-of-the-art results, First three methods use CNN as backend, while the last two use Capsule Net as backend.}
\label{table:results_audio}
\centering
\footnotesize
\begin{tabular}{|>{\centering\arraybackslash}m{2.8cm}|>{\centering\arraybackslash}m{1.5cm}|>{\centering\arraybackslash}m{1.5cm}|} 
\hline
Implementation & Accuracy\\
\hline
\hline
De \textit{et al}~\cite{de2018neural}&94.1\%\\
% \hline
Kim \textit{et al}~\cite{kim2019comparison}&94.82\%\\
\hline
Bae \textit{et al}~\cite{bae2018end}&89.5\%\\
% \hline
\textbf{TimeCaps}& \textbf{92.65\%}\\
% \hline
\hline
\end{tabular}
\vspace{-3mm}
\end{table}
% With these adjustments, TimeCaps was able to achieve 92.65\% accuracy on  Google Speech Commands Dataset surpassing the state-of-the-art capsule network accuracy 89.5\% \cite{bae2018end}.
% \vspace{-3mm}
% \begin{figure}[!h]
% % \centering
% \subfigure[Weights learnt by the $\mathit{concatenation\_layer}$ for the ECG dataset.\label{fig:weight_ecg}]{\includegraphics[scale=0.285]{fig/alpha_beta.pdf}}\hfill
% \subfigure[Weights learnt by the $\mathit{concatenation\_layer}$ for the audio dataset. \label{audio_ab}]{\includegraphics[scale=0.285]{fig/alpha_beta_audio.pdf}}
% \caption{$\alpha$ corresponds to the contribution from TimeCaps Cell $\mathit{A}$ and $\beta$ is corresponds to the contribution from TimeCaps Cell $\mathit{B}$}
% % \label{fig:audio_ab}
% \end{figure}
%Suranga - Here you might need to give some examples on what the adjustments you did

% \begin{figure}[!h]
% \centering
% \includegraphics[scale=0.6]{fig/alpha_beta_audio.pdf}
% \caption{Weights learnt by the $\mathit{concatenation\_layer}$ for the audio dataset. $\alpha$ corresponds to the contribution from TimeCaps Cell $\mathit{A}$ and $\beta$ is corresponds to the contribution from TimeCaps Cell $\mathit{B}$}
% \label{fig:audio_ab}
% \end{figure}
% \vspace{-2mm}
\subsection{Reconstruction Results}
% \subsection{Pertabation Results}
\subsubsection{Effect of each instantiation parameter on Reconstruction Signal}
Each of the instantiation parameter learns a temporal property of the signal. Therefore, we were able to successfully parameterize each ECG signal of length 360 samples using 8 independent parameters. It was observed that, higher the number of parameters, lower the unique temporal relationships captured by one parameter, and thus, higher the correlations among instantiation parameters. Hence, with a sufficient number of instantiation parameters, it is possible to completely characterize each and every temporal property present in the 1D signal uniquely.
% \textcolor{blue}{Athif: I do not udnerstand this sentence starting from  it was observed that... to.. individually}.
This demonstrated capability of data driven parametrization of time series signals proves to be relevant in applications in physiological signals like ECG, specially because of the very high variability between individuals and the existence of rare and hitherto unknown patterns which render conventional parametric modelling less successful. 

Due to the class independent decoder, instantiation parameter vector for the whole dataset is distributed in the same space allowing each instantiation parameter to have similar effect on the reconstruction signal regardless of the class. For example, for normal class, $p1$ parameter corresponds to a change inf R peak amplitude (as illustrated in Fig. \ref{fig:recon_normal}) and the same instantiation parameter $p1$ causes a change in R peak amplitude of other classes as well. Since each input signal can be parameterized using TimeCaps, different aspects of the signal can be modified by altering the value of the respective instantiation parameter. see Fig. \ref{fig:purturb}. 

\newcommand{\W}[1]{\parbox[c]{0.01cm}{\includegraphics[width=1.19cm]{#1}}}
\begin{figure}[h]
\centering
\footnotesize
\begin{tabular}{|>{\hspace{-5pt}}p{0.81cm}|>{\hspace{-5pt}}p{0.81cm}|>{\hspace{-5pt}}p{0.81cm}|>{\hspace{-5pt}}p{0.81cm}|>{\hspace{-5pt}}p{0.81cm}|>{\hspace{-5pt}}p{0.81cm}|>{\hspace{-5pt}}p{0.81cm}|} 
% \begin{tabular}{|p{1.5cm}|c|}
 \hline
\W{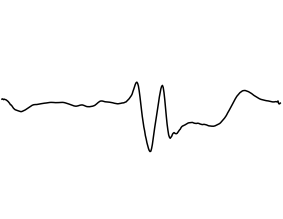}& 
\W{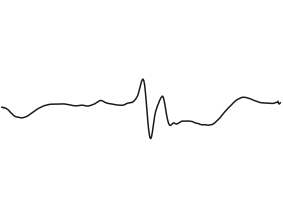}&
\W{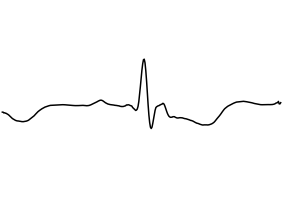}&
\W{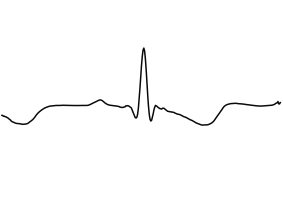}&
\W{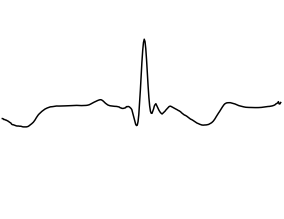}& 
\W{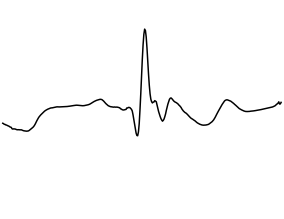}&
\W{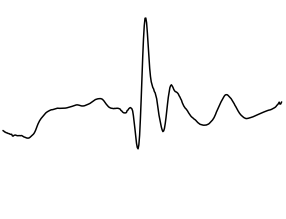}\\
\hline
\W{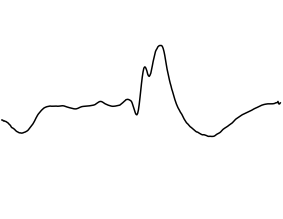}& 
\W{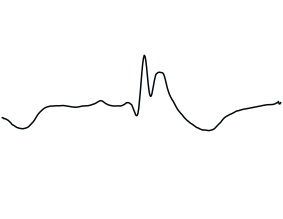}&
\W{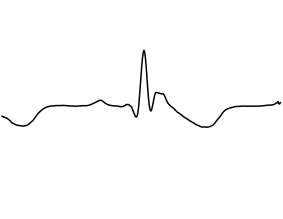}&
\W{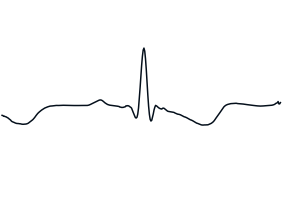}&
\W{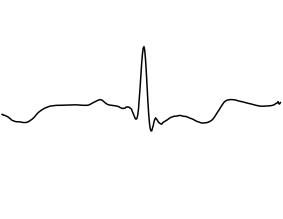}& 
\W{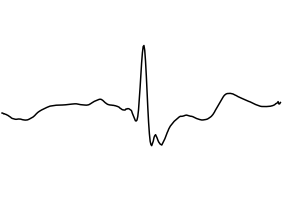}&
\W{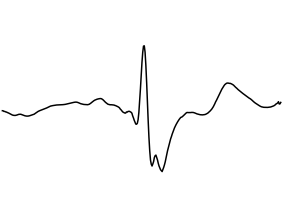}\\
\hline
\W{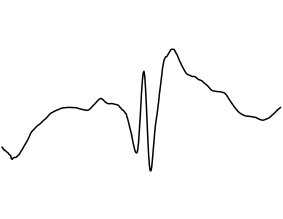}& 
\W{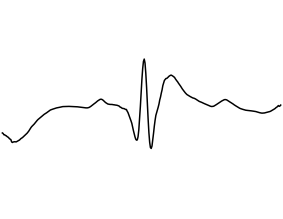}&
\W{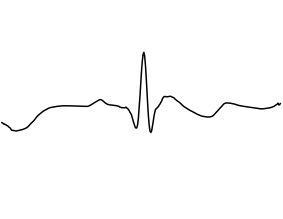}&
\W{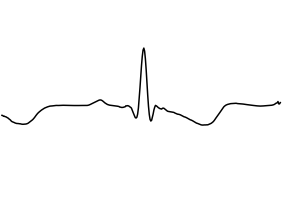}&
\W{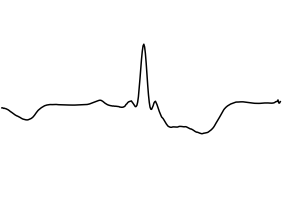}& 
\W{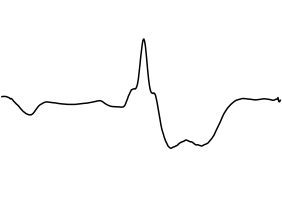}&
\W{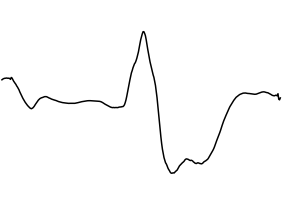}\\
\hline
\W{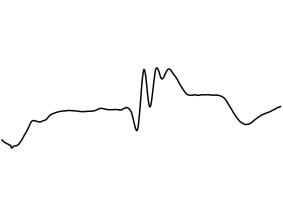}& 
\W{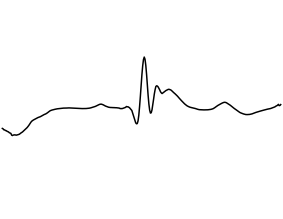}&
\W{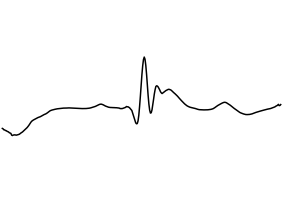}&
\W{2237_1.png}&
\W{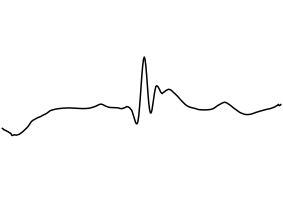}& 
\W{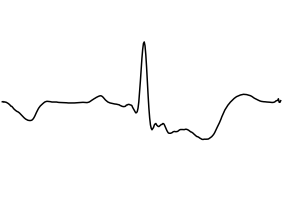}&
\W{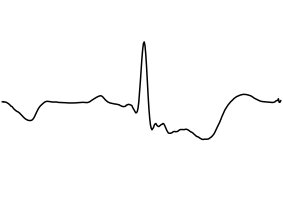}\\
\hline
\W{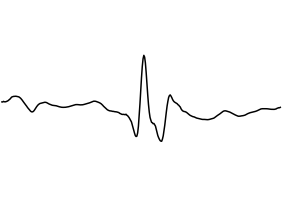}& 
\W{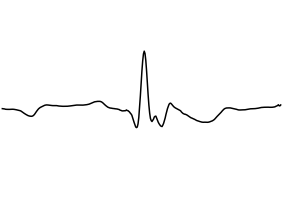}&
\W{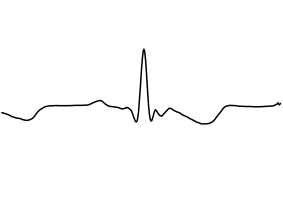}&
\W{2237_1.png}&
\W{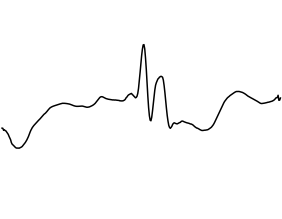}& 
\W{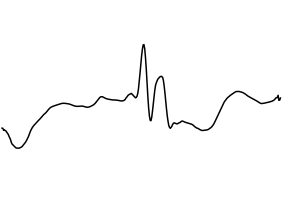}&
\W{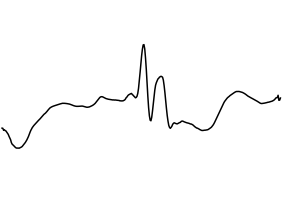}\\
\hline
\end{tabular}

\caption{Perturbation on single normal beat. Centre column is unperturbed. Columns 3, 2 and 1 are perturbed by adding -0.2, -0.5 and -1 while columns 5, 6 and 7 are perturbed by adding 0.2, 0.5 and 1 respectively in the 4 instantiation parameters represented in rows. The first three rows represent perturbations on $p1,p2,p3$ and $p4$ which cause significant morphological changes to the signal while the last row represent the perturbations on $p6$ which does not produce clinically significant morphological changes on the reconstructed signal.}
\label{fig:recon_normal}
\vspace{-2mm}
\end{figure}

We have tabulated the features corresponding to the each instantiation parameter in Table \ref{table:inst_para}, and we observed that $p0$ and $p3$ always affect the amplitude of the T wave while $p1$ was responsible for the R peak amplitude. TimeCaps has enabled data-driven quantitative characterization of different arrhythmia represented by variations in these diagnostic components of the ECG signal. 
Several examples are given in Fig.\ref{fig:purturb}.

\begin{figure}[!h]
  \centering
  \includegraphics[angle=-90, scale=0.45]{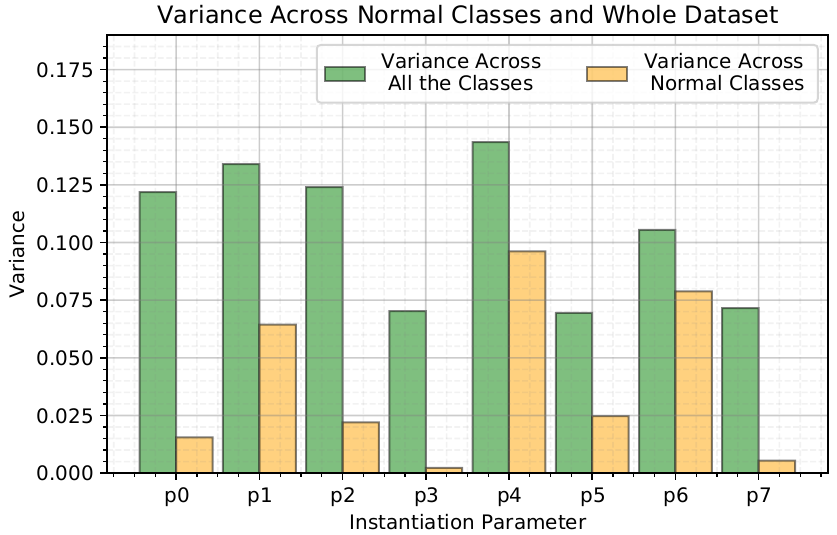}  %\setlength{\belowcaptionskip}{-1cm}
  \caption{All the 8 instantiation parameters and its variance across the MIT-BIH dataset and Normal class of the MIT-BIH dataset.}
 \label{fig:variance}
 \vspace{-2mm}
\end{figure}

\newcommand{\U}[1]{\parbox[c]{0.05cm}{\includegraphics[width=1.19cm]{#1}}}
\begin{figure}[!h]
\centering
\footnotesize
\vspace{-2mm}
\begin{tabular}{|>{\hspace{-5pt}}p{0.81cm}|>{\hspace{-5pt}}p{0.81cm}|>{\hspace{-5pt}}p{0.81cm}|>{\hspace{-5pt}}p{0.81cm}|>{\hspace{-5pt}}p{0.81cm}|>{\hspace{-5pt}}p{0.81cm}|>{\hspace{-5pt}}p{0.81cm}|} 
% \begin{tabular}{|p{1.5cm}|c|}
 \hline
\U{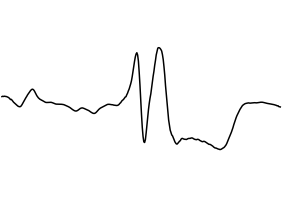}& 
\U{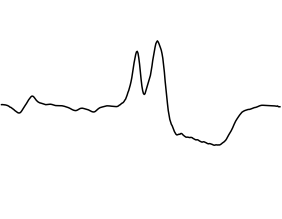}&
\U{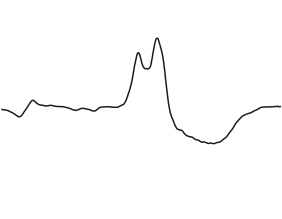}&
\U{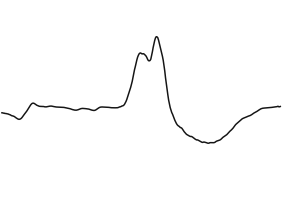}&
\U{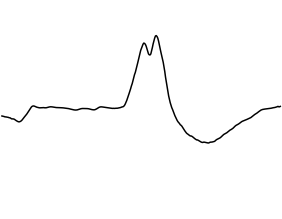}& 
\U{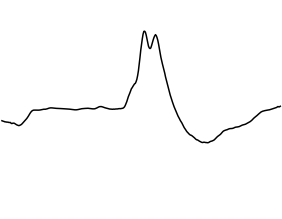}&
\U{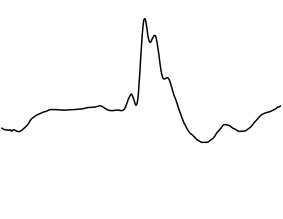}\\
\hline
\U{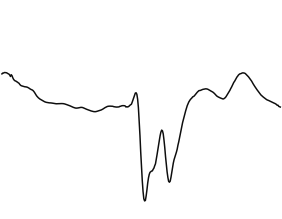}& 
\U{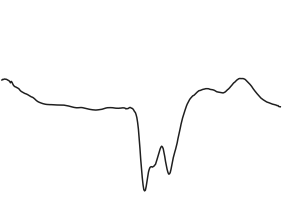}&
\U{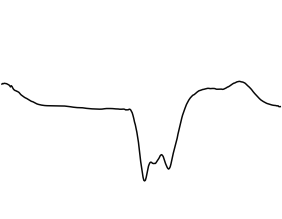}&
\U{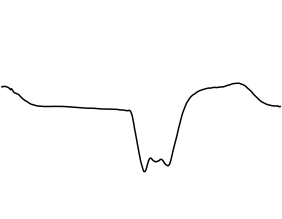}&
\U{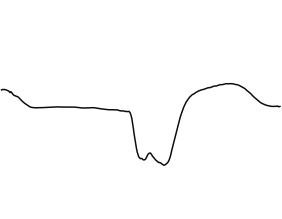}& 
\U{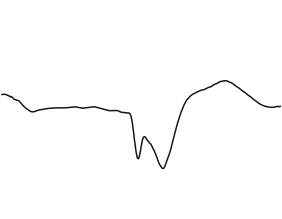}&
\U{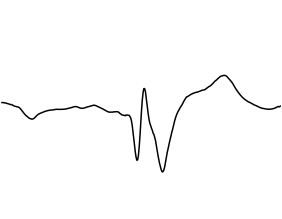}\\
\hline
\U{2237_1-1.png}& 
\U{2237_1-05.png}&
\U{2237_1-02.png}&
\U{2237_1.png}&
\U{2237_1+02.png}& 
\U{2237_1+05.png}&
\U{2237_1+1.png}\\
\hline
\U{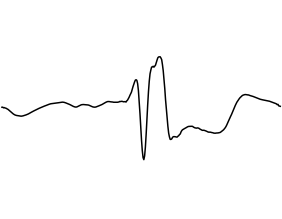}& 
\U{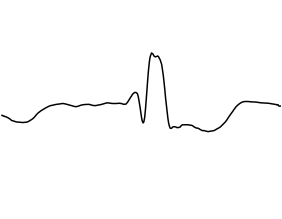}&
\U{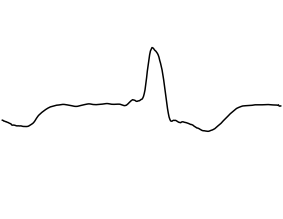}&
\U{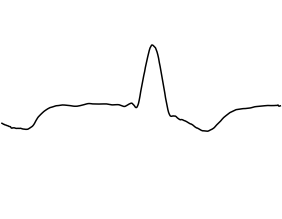}&
\U{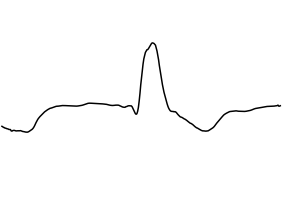}& 
\U{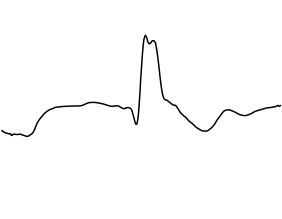}&
\U{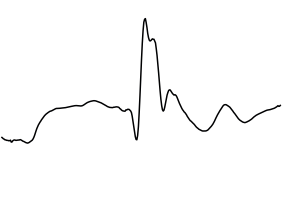}\\
\hline
% \U{3106_1-1.png}& 
% \U{3106_1-05.png}&
% \U{3106_1-02.png}&
% \U{3106_1.png}&
% \U{3106_1+02.png}& 
% \U{3106_1+05.png}&
% \U{3106_1+1.png}\\
% \hline
\end{tabular}
\caption{Signals generated by our decoder network. Each row represents a different class of ECG rhythms. Centre column is unperturbed. Columns 3, 2 and 1 are perturbed by adding -0.2, -0.5 and -1 while columns 5, 6 and 7 are perturbed by adding 0.2, 0.5 and 1 respectively to the $p1$ instantiation parameter. We can observe that, irrespective of the class, $p1$ makes the similar perturbations across all.}
\vspace{-5mm}
\label{fig:purturb}
\end{figure}
% \begin{figure}[!h]
%   \centering
%   \includegraphics[scale=0.4]{fig/variance.png}  %\setlength{\belowcaptionskip}{-1cm}
%   \caption{All the 8 instantiation parameters and its variance across the MIT-BIH dataset and Normal class of the MIT-BIH dataset.}
%  \label{fig:variance}
% \end{figure}
When we rank the instantiation parameters by variance %\textcolor{blue}{Athif: variance of what} 
across the MIT-BIH dataset and Normal class of the MIT-BIH dataset as illustrated in Fig. \ref{fig:variance}, we can observe that the instantiation parameters with high variance cause significant morphological changes while parameters with low variance do  not produce clinically significant morphological changes on the reconstructed signal. This strengthen the observations recorded in \ref{table:inst_para} based on the results of ECG dataset.
\begin{table}[!h]
\caption{Effect of each instantiation parameter on reconstruction signal when number of instantiation parameter is 8}
\label{table:inst_para}
\centering
\footnotesize
\begin{tabular}{|>{\centering\arraybackslash}m{0.5cm}|>{\centering\arraybackslash}m{2.8cm}|>{\centering\arraybackslash}m{2.8cm}|>{\centering\arraybackslash}m{0.9cm}|} 
\hline
Inst. Para. & Change with increasing perturbation value  &  Change with decreasing perturbation value& ECG significance\\
\hline
p0&T wave amplitude decreases&T wave amplitude increases&Strong\\ 
\hline
p1&R peak amplitude increases&R peak amplitude decreases &Strong\\
\hline
p2&Q peak amplitude decreases&Q peak amplitude increases &Strong\\
\hline
p3&T wave amplitude increases&T wave amplitude decreases&Strong\\
\hline
p4&ST segment elevated&ST segment depressed&Strong\\ 
\hline
p5&ST segment elevated and rotated&ST segment elevated and rotated&Weak\\
\hline
p6&Overshoot immediately following S dip increases &overshoot immediately following S dip decreases&Weak\\
\hline
p7&No significant and meaningful change noticed&No significant and meaningful change noticed&Weak\\
\hline
\end{tabular}
\vspace{-1mm}
\end{table}
\subsubsection{Effect of Different number of Instatntiation Parameters}
\newcommand{\V}[1]{\parbox[c]{0.05cm}{\includegraphics[width=2.5cm]{#1}}}
\begin{figure}[!h]
\vspace{-2mm}
\centering
\footnotesize
\begin{tabular}{|>{\hspace{-5pt}}p{0.77cm}|>{\hspace{-5pt}}p{2.12cm}|>{\hspace{-5pt}}p{2.12cm}|>{\hspace{-5pt}}p{2.12cm}|} 
% \begin{tabular}{|p{1.5cm}|c|}
 \hline
Original Signal& 
\V{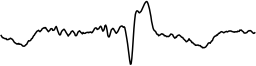}&
\V{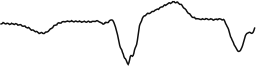}&
\V{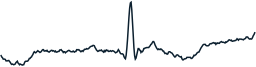}\\
\hline
Recon 4&
\V{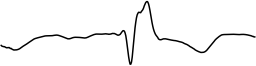}&
\V{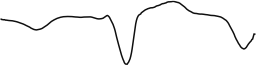}&
\V{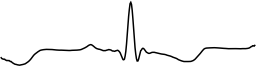}\\
\hline
Recon 8&
\V{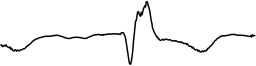}&
\V{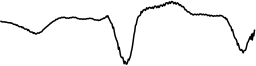}&
\V{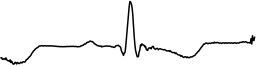}\\
\hline
Recon 16&
\V{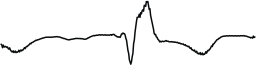} &
\V{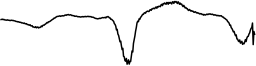} &
\V{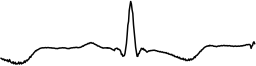}\\
\hline
Recon 24&
\V{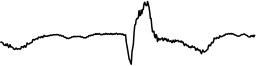}&
\V{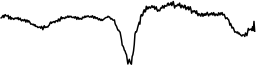}&
\V{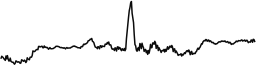}\\
\hline
% \end{tabular}
% \vspace{-2mm}
% \end{figure}
% \begin{figure}[H]
% \centering
% \footnotesize
% \begin{tabular}{|>{\hspace{-5pt}}p{0.77cm}|>{\hspace{-5pt}}p{2.12cm}|>{\hspace{-5pt}}p{2.12cm}|>{\hspace{-5pt}}p{2.12cm}|} 
% \begin{tabular}{|p{1.5cm}|c|}
\multicolumn{4}{c}{  } \\
\hline
Original Signal& 
\V{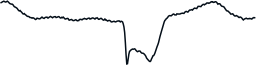}&
\V{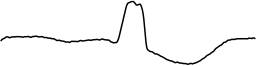}&
\V{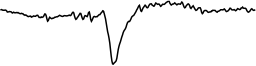}\\
\hline
Recon 4&
\V{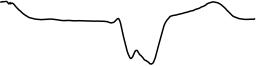}&
\V{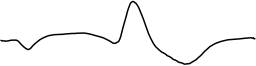}&
\V{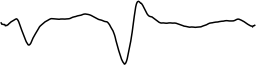}\\
\hline
Recon 8&
\V{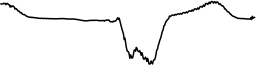}&
\V{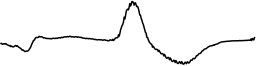}&
\V{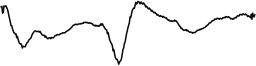}\\
\hline
Recon 16&
\V{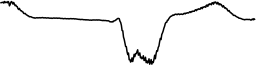} &
\V{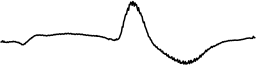} &
\V{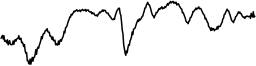} \\
\hline
Recon 24&
\V{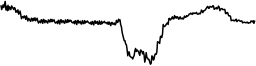}&
\V{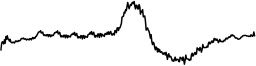}&
\V{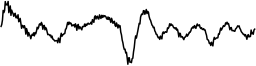}\\
\hline
\end{tabular}
\caption{Reconstruction results for different instantiation parameters. We investigate the quality of the reconstruction with the variation of the number of instantiation parameters.}
\label{fig:recon}
\vspace{-6mm}
\end{figure}
Decoder network has jointly learnt to decode the encoded signal which was parameterized by the TimeCaps network. Fig.~\ref{fig:recon} shows sample of reconstructed signals.

Further, Table~\ref{table:inst_para_numbers} demonstrates the effect of having low number of instantiation parameters at the $\mathit{classification\_capsules}$ layer. When the number of instantiation parameters decreases, the $\mathit{mean\_square\_error}$ between the input signal and the reconstructed signal also decreases. Yet, as illustrated in Fig. \ref{fig:recon}, the reconstructed signal can be employed in applications which are required to perform well with low quality signals.    

\begin{table}[!h]
\vspace{-2mm}
\caption{Comparison of different number of instantiation parameters}
\label{table:inst_para_numbers}
\centering
\footnotesize
\begin{tabular}{|>{\centering\arraybackslash}m{1.5cm}|>{\centering\arraybackslash}m{1.5cm}|>{\centering\arraybackslash}m{1cm}|>{\centering\arraybackslash}m{1.3cm}|} 
\hline
Number of Inst. Parameters & MSE loss at decoder  &  Accuracy&Inference Time sec/sample\\
\hline
4&0.0149&95.99\%&0.004\\
\hline
8&0.0137&96.12\%&0.004\\
\hline
16&0.0318&96.21\%&0.005\\ 
\hline
24&0.0132&96.08\%&0.005\\
\hline
\end{tabular}
\vspace{-2mm}
\end{table}
\noindent Further, when the number of instantiation parameters increases, it introduces unnecessary artifacts to the reconstruction signal as illustrated in \textbf{Recon 24}.
% \vspace{-1mm}
\section{Conclusion}
\label{sec:conclusion}
This paper introduces a novel CapsuleNetwork based architecture, TimeCaps which was tailored to classify, decode, and encode ECG signals. Our results indicated that TimeCaps performs on par with other state of the art methods. Also, one major advantage of TimeCaps is its ability to reconstruct raw ECG with very low dimensional latent representation. We also evaluated the performance of TimeCaps in a raw audio classification task to evaluate its performance on other types of time series data. Results indicated that we can surpass the state-of-the-art results with capsule network.

As future work, capabilities of TimeCaps can be further extended to synthesize ECG data samples which might be helpful in improving the accuracy of rare beat types. Possible methods include, adding a controlled noise to the instantiating parameters as suggested by Sabour \textit{et al}.\cite{sabour2017dynamic}.

% \vspace{-1mm}
\bibliographystyle{IEEEtran}
\bibliography{main}
\end{document}